\documentclass[letterpaper, 10pt, journal, twoside]{classes/IEEEtran}
\IEEEoverridecommandlockouts

\usepackage{packages/decar-common}
\usepackage{packages/decar-dynamics}
\usepackage{packages/decar-lie}
\usepackage{packages/decar-post} 

\usepackage{caption}
\usepackage{booktabs}
\usepackage[T1]{fontenc}
\usepackage{titlesec}
\titlespacing*{\subsection}{0.3em}{0.3em}{0.3em}
\titlespacing*{\section}{0.2em}{0.2em}{0.2em}

\usepackage[disable]{todonotes}

\newcommand{\wkcomment}[2][] {\todo[inline,backgroundcolor=red!20!white, #1]{(Wouter) #2}}

\newsavebox{\subfigbox}

\makeatletter
\AtBeginDocument{
  \check@mathfonts
}
\makeatother

\graphicspath{{figs/}}

\renewcommand{\expect}[2]{\mathbb{E}_{#1}\left[#2\right]}
\def\-{\text{-}}
\def\+{\text{+}}

\title{\LARGE \bf
    Gaussian Variational Inference with Non-Gaussian Factors for State Estimation: A UWB Localization Case Study
}

\author{
    Andrew Stirling$^{1}$,
    \and
    Mykola Lukashchuk$^{2}$,
    \and
    Dmitry Bagaev$^{2}$,
    \and
    Wouter Kouw$^{2}$,
    \and
    James R. Forbes$^{1}$
    \thanks{Manuscript received: August, 31, 2025; Revised: November, 18, 2025; Accepted: December, 9, 2025.}
    \thanks{This paper was recommended for publication by Editor Sven Behnke upon evaluation of the Associate Editor and Reviewers' comments. This work was supported by the Natural Sciences and Engineering Research Council of Canada (NSERC) Alliance Grant program, Denso Corporation, ARA Robotics, and the NSERC-FRQ NOVA program.}
    \thanks{$^{1}$Andrew Stirling and James R. Forbes are with the Department of Mechanical Engineering, McGill University, Canada ({\scriptsize e-mail: \tt andrew.stirling2@mail.mcgill.ca, james.richard.forbes@mcgill.ca}).}
    \thanks{$^{2}$Mykola Lukashchuk, Dmitry Bagaev and Wouter Kouw are with the Department of Electrical Engineering, TU Eindhoven, Netherlands ({\scriptsize e-mail: \tt \{m.lukashchuk, d.v.bagaev, w.m.kouw\}@tue.nl}).}
    \thanks{Digital Object Identifier (DOI): see top of this page.}
}

\begin{document}
%
%
%
%
%
%
%
\def \myJournal {IEEE Robotics and Automation Letters}
\def \myDoi {10.1109/LRA.2026.3653370}
\def \myPaperSiteName {IEEE Xplore}
\def \myPaperSiteLink {https://ieeexplore.ieee.org/document/11347486}
\def \myYear {2026}
\def \myPaperCitation{A. Stirling, M. Lukashchuk, D. Bagaev, W. Kouw and J. R. Forbes, "Gaussian Variational Inference with Non-Gaussian Factors for State Estimation: A UWB Localization Case Study," in IEEE Robotics and Automation Letters, Jan 2026}


\begin{figure*}[t]

\thispagestyle{empty}
\begin{center}
\begin{minipage}{6in}
\centering
This paper has been accepted for publication in \emph{\myJournal}. 
\vspace{1em}

This is the author's version of an article that has, or will be, published in this journal or conference. Changes were, or will be, made to this version by the publisher prior to publication.
\vspace{2em}

\begin{tabular}{rl}
DOI: & \myDoi\\
\myPaperSiteName: & \texttt{\myPaperSiteLink}
\end{tabular}

\vspace{2em}
Please cite this paper as:

\myPaperCitation

\vspace{15cm}
\copyright \myYear \hspace{4pt}IEEE. Personal use of this material is permitted. Permission from IEEE must be obtained for all other uses, in any current or future media, including reprinting/republishing this material for advertising or promotional purposes, creating new collective works, for resale or redistribution to servers or lists, or reuse of any copyrighted component of this work in other works.

\end{minipage}
\end{center}
\end{figure*}
\newpage
\clearpage
\pagenumbering{arabic} 

\maketitle
\vspace{-2em}
\markboth{IEEE Robotics and Automation Letters. Preprint Version. Accepted Dec, 2025}
{Stirling \MakeLowercase{\textit{et al.}}: Gaussian Variational Inference with Non-Gaussian Factors for State Estimation} 

\fontdimen16\textfont2=\fontdimen17\textfont2
\fontdimen13\textfont2=5pt


\begin{abstract}
	This letter extends the exactly sparse Gaussian variational inference (ESGVI) algorithm for state estimation in two complementary directions. First, ESGVI is generalized to operate on matrix Lie groups, enabling the estimation of states with orientation components while respecting the underlying group structure. Second, factors are introduced to accommodate heavy-tailed and skewed noise distributions, as commonly encountered in ultra-wideband (UWB) localization due to non-line-of-sight (NLOS) and multipath effects. Both extensions are shown to integrate naturally within the ESGVI framework while preserving its sparse and derivative-free structure. The proposed approach is validated in a UWB localization experiment with NLOS-rich measurements, demonstrating improved accuracy and comparable consistency. Finally, a \textsc{Python} implementation within a factor-graph-based estimation framework is made open-source to support broader research use.
\end{abstract}



\raggedbottom
\section{Introduction}
\label{sec:introduction}
\IEEEPARstart{A}{utonomous} robots must estimate their position, orientation, and other key pieces of information to operate effectively. This collection of dynamic variables, known as the state of the robot, fully describes the system at a given time. Consistent and accurate state estimation is essential for high-level decision-making and the support of downstream tasks such as guidance and control. Within state estimation, approaches to localization can be broadly split into the filtering and batch approaches \cite{Barfoot_2024}. Filtering methods use data up to the current timestep to estimate the robot’s state, whereas batch methods use all available data over a given time period. The latter forms the foundation of this work. Taking a Bayesian perspective, the goal of state estimation is to compute the state $\mbf{x} \in \rnums^{n_x}$, given a set of measurements $\mbf{y} \in \rnums^{n_y}$, that is, the full posterior $p(\mbf{x} | \mbf{y})$ is computed, given a prior $p(\mbf{x})$. Using Bayes' rule, the relationship
\begin{align}
    p(\mbf{x} | \mbf{y}) = \frac{p(\mbf{y} | \mbf{x}) p(\mbf{x})}{p(\mbf{y})} = \frac{p(\mbf{x}, \mbf{y})}{p(\mbf{y})},
\end{align}
is exploited to compute the posterior. For nonlinear measurement models $p(\mbf{y} | \mbf{x})$, the posterior is \emph{not} a Gaussian probability density function (PDF). Currently, the majority of the batch state estimation literature is based on solving a \emph{maximum-a-posteriori} (MAP) problem, which involves computing the maximum, that is the mode, of the Bayesian posterior. An alternative to MAP is variational inference \cite{Bishop_2006}, which approximates the posterior by fitting a distribution that minimizes the Kullback–Leibler (KL) divergence. In \cite{esgvi_paper}, exactly sparse Gaussian variational inference (ESGVI) was introduced as a tractable method for computing Gaussian approximations of the posterior in large-scale nonlinear batch state estimation problems, by exploiting the sparsity associated with robotics problems. By accounting for the full shape of the posterior distribution, ESGVI was shown to outperform MAP in nonlinear state estimation problems. However, in its current form, ESGVI operates on vector-valued states and, although the use of a robust cost function was suggested as a future line of inquiry in~\cite{esgvi_paper}, it has only been evaluated with Gaussian-distributed sensor models. In practice, sensor measurements can deviate significantly from Gaussian assumptions, producing outliers that can severely degrade estimation performance. This issue is particularly evident in ultra-wideband (UWB) localization, where ranging measurements are increasingly used in robotics due to their long-range and low-cost characteristics \cite{Shalaby_Cossette_Forbes_Le_Ny_2023}. Nevertheless, UWB measurements are often corrupted by non-line-of-sight (NLOS) effects and multipath propagation, leading to heavy-tailed and asymmetric noise distributions that must be accounted for \cite{Kim_Yoon_Lee_2024}.
To address these limitations, ESGVI is extended to operate on matrix Lie group (MLG) states, a modest but necessary step, as MLGs naturally capture the geometry of states commonly found in robot navigation problems \cite{Solà_Deray_Atchuthan_2021}. In addition, non-Gaussian measurement factors are incorporated, providing a more flexible way to model heavy-tailed and skewed noise distributions. While such factors have been explored in related estimation frameworks~\cite{Vilà-Valls_Vincent_Closas_2019}, the contribution here lies in demonstrating their integration within the ESGVI setting and evaluating their impact in practice. Although the proposed extensions are general, their effectiveness is illustrated through a specific UWB example for concreteness and ease of interpretation. Building on~\cite{esgvi_paper}, the contributions of this letter are as follows:
\begin{itemize}
    \item Extending ESGVI to support state estimation on matrix Lie groups.
    \item Introducing non-Gaussian measurement factors into the ESGVI framework, tailored for asymmetric and skewed noise distributions.
    \item Evaluating ESGVI in a real-world, outlier rich, UWB localization experiment, demonstrating greater accuracy and comparable consistency.
    \item Releasing an open-source \textsc{Python} implementation of ESGVI within a factor-graph-based estimation framework at https://github.com/decargroup/gvi{\textunderscore}ws.
\end{itemize}

In this letter bold uppercase letters such as $\mbf{A}$ denote matrices, bold lowercase letters $\mbf{a} \in \rnums^{n}$ denote column vectors, while lowercase letters $a \in \rnums$ represent scalar values.

\section{Problem Statement}\label{sec:problem}
\begin{figure}[htbp]
    \centering
    \begin{subfigure}{\columnwidth}
        \centering
        \includegraphics[width=\linewidth]{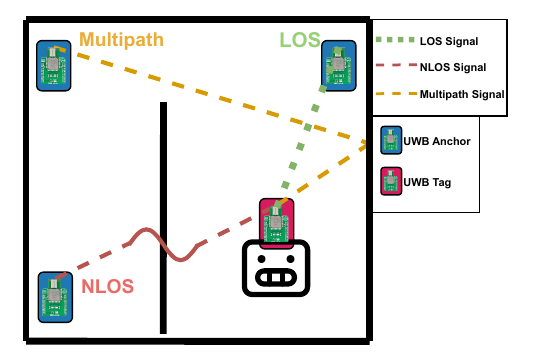}
        \caption{Illustration of LOS, NLOS and multipath signal propagation between fixed UWB anchors and a mobile UWB tag attached to the robot.}
        \label{fig:nlos_scenario}
    \end{subfigure}
    
    \vspace{-0.2em} 
    
    \begin{subfigure}{\columnwidth}
        \centering
        \includegraphics[width=\linewidth]{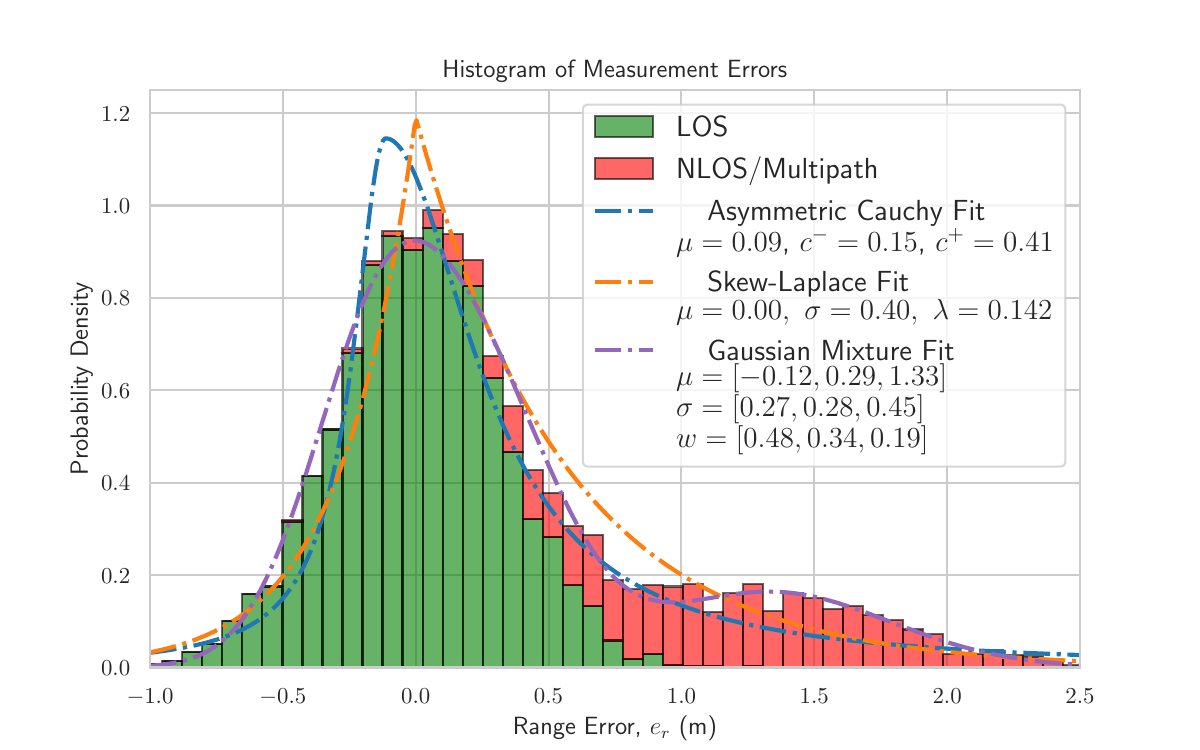}
        \caption{Simulated noise distribution modeling NLOS and multipath effects, where 25\% of Gaussian range measurements are corrupted with values drawn uniformly between 1 and 6 standard deviations. Various probabilistic models are fitted and compared for estimation performance in Section~\ref{sec:sim_results}.}
        \label{fig:uwb_fit_sim}
    \end{subfigure}
    \caption{Depiction of obstacle-induced signal propagation effects that lead to errors in range-based localization.}
    \label{fig:problem}
    \vspace{-1em}
\end{figure}

Consider a robot in an indoor environment estimating its state using UWB range measurements $y_r$, from a set of fixed UWB anchors positioned throughout the room. When the signal path is unobstructed and in line-of-sight (LOS), the range error $e_r$, defined as the difference between the measured and true distance, is typically small. In contrast, when obstacles block the direct path, creating non-line-of-sight (NLOS) conditions, signals may reflect from surrounding surfaces, or even pass through obstacles. These multipath and NLOS measurements bias the ranges resulting in a skewed error distribution, as shown in Figure~\ref{fig:problem}.

A second challenge arises from the fact that some state components, such as position, admit a natural Euclidean representation, whereas others are inherently directional. For example, the vehicle’s orientation is periodic, with $0$~$(\si{rad})$ and $2\pi$~$(\si{rad})$ denoting the same heading. Treating rotational variables in Euclidean space $\rnums^{n_x}$\;neglects their non-commutative group structure \cite{Barfoot_Furgale_2014}, leading to singularities and distorted uncertainty representations~\cite{long2013banana}. To address this, the vehicle state is represented as a matrix Lie group (MLG), that is, $\mbf{X}_k \in \SE{2}$, which combines both translational and rotational components as~\cite{Solà_Deray_Atchuthan_2021,Barfoot_2024},
\begin{equation}
    \label{eq:state_representation}
    \mbf{X}_k = 
    \bbm 
    \mbf{C}_{ab} & \mbf{r}^{zw}_a \\
    \mbf{0} & 1
    \ebm \in \SE{2},
\end{equation}
where $\mbf{C}_{ab} \in \SO{2}$ is the rotation matrix mapping the body frame $\rframe{b}$ to the absolute frame $\rframe{a}$. The notation $\mbf{r}^{zw}_a$ represents the position of the point $z$ on the robot relative to the origin $w$ of the absolute frame, resolved in the frame $\rframe{a}$. A broader overview of the required MLG operations used throughout this work is provided in Appendix~\ref{sec:appendix:lie}. By representing the state as a MLG, the underlying manifold structure is respected, and directional uncertainty is properly captured. The objective is to leverage this representation to estimate the robot state as accurately as possible under skewed measurement noise conditions. Standard estimation approaches assume Gaussian sensor likelihoods and thus struggle in this setting, often leading to bias or even numerical instability. A robust solution must instead incorporate skewed likelihood functions while still producing a Gaussian distribution for the state estimate.
\section{Background and Related Work}

\subsection{Probabilistic Model Specification}
\label{sec:state_estimation:model}

This work focuses on state-space models, that is, models that contain explicit dynamics and observation functions. For simplicity, consider the vector-valued state $\mbf{x}_k$ at time index $k=0$, with an initial state distribution of
\begin{align}
    p(\mbf{x}_0) = \mc{N}(\mbf{x}_0 \mid \mbs{\mu}_0, \mbs{\Sigma}_0) \, .
\end{align}
While attention is restricted to vector states here for clarity, in later sections this framework will be generalized to accommodate the MLG state, $\mbf{X}_k$.
The state evolves over time according to a stochastic state transition with the Markov property, i.e., only dependent on the previous state \cite[Property 4.1]{Sarkka}. It is modelled as a Gaussian distribution,
\begin{align}
    p(\mbf{x}_{k} \mid \mbf{x}_{k-1}, \mbf{u}_{k-1}) = \mathcal{N}(\mbf{x}_{k} \mid \mbf{f}(\mbf{x}_{k-1}, \mbf{u}_{k-1}), \mbf{Q}_k) \, ,
\end{align}
where $\mbf{u}_{k-1}$ represents the input applied at the previous step, $\mbf{f}(\cdot)$ is the nonlinear dynamics function, and $\mbf{Q}_k$ is the discrete-time process noise covariance. We implicitly assume that stochastic perturbations to the state are independent over time.
For the sensor measurement likelihood, non-Gaussian likelihoods can be incorporated when the noise characteristics are known. In particular, for range measurements $y_k$, the likelihood is modeled using a Skew-Laplace distribution of the form \cite{Vilà-Valls_Vincent_Closas_2019}, 
\begin{align} \label{eq:model:Skew-Laplace}
    p(y_k &| \mbf{x}_k) = \mc{SL}(y_k | g(\mbf{x}_k), \sigma, \lambda), \\
    \label{eq:model:Skew-Laplace:pdf}
    &= \frac{1}{2\sigma \alpha} \exp \left(\lambda \left(\frac{y_k \! - \! g(\mbf{x}_k)}{\sigma^2}\right)- \alpha \left|\frac{y_k \! - \! g(\mbf{x}_k)}{\sigma}\right|\right) \, ,
\end{align}
where $g(\cdot)$ denotes the predicted range as a function of the state, and the parameters, $\sigma \in \rnums^{+}, \, \lambda \in \rnums$, denoting the distribution scale, and skewness, where $\alpha = \sqrt{1 + (\lambda / \sigma)^2}$. The parameters are estimated separately using maximum likelihood. 
Combining the prior, process, and measurement models together, the probabilistic model of the discrete-time series of $k = 1, \ldots, K$ is,
\begin{align} \label{eq:model_joint}
    p(\mbf{y}, \mbf{x} | \mbf{u}) = p(\mbf{x}_0) \prod_{k=1}^{K} p(\mbf{x}_k | \mbf{x}_{k-1}, \mbf{u}_{k-1}) p(y_k | \mbf{x}_k) \, .
\end{align}
\wkcomment{You forgot to include $\mbf{u}$ in $p(\mbf{x}, \mbf{z})$. This should still be updated throughout the rest of the paper.\\

I see that the rest of the paper uses $\mbf{z}$. I can define $\mbf{z} = \{(y_k, u_k)\}_{i=0}^T$, then most equations below are again consistent.
}
The next section discusses the proposed method for obtaining the marginal posterior over the trajectory $\mbf{x} = \{ \mbf{x}_0, \ldots, \mbf{x}_K \}$.
\subsection{State of the Art}
A common approach to batch state estimation is to assume a Gaussian measurement model and find the most probable state vector under the state posterior distribution (i.e., MAP estimation),
\begin{align} \label{eq:inference:MAP}
    \hat{\mbf{x}} &= \underset{\mbf{x}}{\arg \max} \ p(\mbf{x} | \mbf{y}, \mbf{u}).
\end{align}
By assuming Gaussian noise and minimizing the negative logarithm of the posterior, the MAP estimate can be formulated as a nonlinear least-squares (NLS) problem~\cite[\S3.1.2]{Barfoot_2024}. Expressed in this sense, the estimator minimizes a sum-of-squared-errors cost function, which makes it highly sensitive to outliers~\cite[\S5.4]{Barfoot_2024}. To address the sensitivity of MAP-based NLS estimators to outliers, a variety of strategies have been explored, particularly in UWB-based state estimation. For example, Omni-Swarm~\cite{omni_swarm} introduced a visual–inertial–UWB estimation pipeline formulated as a MAP-based NLS optimization problem, where UWB outliers were rejected using fixed elevation-angle criteria. Distributed approaches~\cite{Fishberg_How_2022,murp, vir_slam} extend this concept to multi-agent settings, also solving MAP NLS problems while rejecting inconsistent range measurements through averaging, geometric or max velocity constraints. More recently, learning-based methods have been employed to detect and discard NLOS UWB measurements. In~\cite{Qi_Zhao_Guo_Kaliuzhnyi_Wang_2025} and~\cite{Jiang_Shen_Chen_Chen_Liu_Bo_2020}, convolutional neural network (CNN) and long short-term memory (LSTM) architectures were proposed to classify and reject suspected NLOS observations. However, since no model can guarantee perfect outlier identification, residual contamination typically persists in the dataset, introducing skewed measurement noise.

To more systematically account for these residual effects, M-estimators incorporate robust cost functions directly within the NLS formulation, allowing the optimization to down-weight the influence of large residuals rather than reject measurements outright. Various robust loss functions have been proposed~\cite{zhang1997parameter}, and in the context of UWB localization, several noise parameterizations have been studied. For example, in~\cite{uvip} a Huber loss was integrated into the NLS optimization, while~\cite{Shalaby_Cossette_Forbes_Le_Ny_2023} proposed a Cauchy loss, however these symmetric losses ignore the asymmetry present in NLOS noise. To capture skewed characteristics,~\cite{Kok_Hol_Schön_2015, Kim_Yoon_Lee_2024} used an asymmetric two-piece distribution. A similar asymmetric Cauchy distribution is used in this letter defined by
\begin{equation}
p(y_k| \mbf{x}_k) =
\begin{cases}
\alpha \frac{1}{1 + \left( \frac{r(y_k, \mathbf{x}_k)}{c^{-}} \right)^2}, & r(\cdot, \cdot) < 0, \\
\alpha \frac{1}{1 + \left( \frac{r(y_k, \mathbf{x}_k)}{c^{+}} \right)^2}, & r(\cdot,\cdot) \ge 0,
\end{cases}
\label{eq:asym_cauchy_pdf}
\end{equation}
where \( r(\cdot, \cdot) \) is the residual, \( c^{-} \) and \( c^{+} \) are scale parameters, and \( \alpha = 2/[\pi(c^{+}+c^{-})] \) normalizes the distribution. The asymmetric Cauchy loss provides outlier robustness while explicitly modeling skewed measurement noise. A secondary flexible alternative for modeling skewed distributions is to employ a Gaussian mixture measurement model~\cite{Zhao_Goudar_Tang_Qiao_Schoellig_2023},
\begin{align} \label{eq:model:gmm-likelihood}
    p(y_k | \mbf{x}_k) = \sum_{j=1} w_j \mathcal{N}(r(y_k, \mbf{x}_k) | \mbs{\psi}_j),
\end{align}
with weights $w_j\ge 0,\ \sum_j w_j=1$ and $\mbs{\psi}_j=(\mu_j,\Sigma_j)$ \cite{Pfeifer_Lange_Protzel_2021}. Component weights and parameters are calibrated separately, or solved for in parallel with the state $\mbf{x}$, in an Expectation Maximization (EM) framework \cite[\S9.3]{Bishop_2006}. The Gaussian mixture measurement model of~\eqref{eq:model:gmm-likelihood} is incorporated into the joint distribution of~\eqref{eq:model_joint}, and MAP estimation~\eqref{eq:inference:MAP} produces a state estimate. Although Gaussian mixture models (GMMs) are flexible enough to capture skewed noise distributions, they require additional parameters and, while incorporation into a NLS framework is possible, it demands careful handling~\cite{Pfeifer_Lange_Protzel_2021}. Moreover, a MAP estimator provides only a point estimate of the state, with uncertainty obtained using an ad-hoc Laplace approximation \cite[\S4.3.1]{Barfoot_2024}. 
Alternatively, while the methods discussed thus far rely on MAP-based optimization, ESGVI offers a fundamentally different probabilistic framework that estimates both the state and its uncertainty directly through variational inference. Performing parameter learning in parallel with ESGVI~\cite{Wong_Yoon_Schoellig_Barfoot_2020} has been shown to improve robustness, as jointly estimating the covariance with an appropriate prior implicitly rejects outliers. While robustness was identified as an open direction in~\cite{esgvi_paper}, measurement models specific to a non-Gaussian distribution, such as the Skew-Laplace in~\eqref{eq:model:Skew-Laplace:pdf} have not yet been studied directly within the ESGVI framework. Consequently, in this work, comparisons are made against robust MAP approaches rather than existing ESGVI methods. In the following, ESGVI is shown to naturally incorporate robust, non-Gaussian measurement models with matrix Lie group states, enabling principled estimation under heavy-tailed or asymmetric noise. This is illustrated using a Skew-Laplace model for UWB localization, capturing NLOS-induced asymmetry.



\section{Gaussian Variational Inference}
\label{sec:gvi}
As introduced previously, an alternative to MAP estimation is Gaussian variational inference (GVI), which fits a Gaussian approximation $q(\mbf{x}) = \mc{N}(\mbs{\mu}, \mbs{\Sigma})$, to the posterior $p(\mbf{x|y,u})$, by minimizing the Kullback-Leibler (KL) divergence \cite{Bishop_2006},
\begin{equation}
    \label{eq:KL_div}
    \textrm{KL}(q||p) = \expect{q}{\ln q(\mbf{x}) - \ln p(\mbf{x}|\mbf{y,u})}.
\end{equation}
Expanding and dropping terms independent of $q(\mbf{x})$, defines the loss functional
\begin{align}
    \label{eq:loss_functional}
    V(q) = \expect{q}{\phi(\mbf{x})} + \onehalf \ln \left(| \mbs{\Sigma}\inv|\right),
\end{align}
where $\phi(\mbf{x}) = -\ln p(\mbf{x, y , u})$ is the negative log joint state and measurement likelihood, and $|\mbs{\Sigma}\inv|$ is the determinant of the information matrix of the variational approximation, $q(\mbf{x})$. In the work of \cite{esgvi_paper}, an exactly sparse Gaussian variational inference (ESGVI) optimization scheme was defined for the vector state $\mbf{x}$, minimizing $V(q)$. The iterative update to the mean $\mbs{\mu}$, and information matrix $\mbs{\Sigma}\inv$, are defined according to
\begin{subequations}\label{eq:gvi_vector_update}
\begin{align}
    \label{eq:gvi_update_vector_1}
    \left(\mbs{\Sigma}\inv\right)^{(i+1)} &= \expect{q^{(i)}}{\pd{^2 \phi(\mbf{x})}{\mbf{x}^{\trans}\p\mbf{x}}}, \\
    \label{eq:gvi_update_vector_2}
    \left(\mbs{\Sigma}^{-1}\right)^{(i+1)}\delta \mbs{\mu} &= - \expect{q^{(i)}}{\pd{\phi(\mbf{x})}{\mbf{x}^{\trans}}}, \\
    \label{eq:gvi_update_vector_3}
    \mbs{\mu}^{(i+1)} &= \mbs{\mu}^{(i)} + \delta \mbs{\mu},
\end{align}
\end{subequations}
where $\delta\mbs{\mu}$ is computed by solving~\eqref{eq:gvi_update_vector_2}. By making use of the factoring properties of the probabilistic model of~\eqref{eq:model_joint}, the expectations can be expressed over the marginal $q_k(\mbf{x}_k)$. Defining the projection, $\mbf{x}_k = \mbf{P}_k \mbf{x}$, the expectations become,
\begin{subequations}
    \begin{align}
        \expect{q}{\phi(\mbf{x})} &=  \sum_{k=1}^{K} \expect{q_k}{\phi_k(\mbf{x}_k)}, \\
        \expect{q}{\pd{\phi(\mbf{x})}{\mbf{x}^{\trans}}} &= \sum_{k=1}^{K}\mbf{P}_k^\trans  \expect{q_k}{\pd{}{\mbf{x}_k^{\trans}}\phi_k (\mbf{x}_k)}, \\
        \expect{q}{\pd{^2\phi(\mbf{x})}{\mbf{x}^{\trans} \p \mbf{x}}} &= \sum_{k=1}^{K} \mbf{P}_k^\trans  \expect{q_k}{\pd{^2}{\mbf{x}_k^{\trans}\p\mbf{x}_k}\phi_k (\mbf{x}_k)} \mbf{P}_k,
    \end{align}
\end{subequations}
which reduces the computation to smaller, factor-specific subsets. Next, using Stein's lemma \cite[\S2.2.16]{Barfoot_2024}, it was demonstrated how taking derivatives of $\phi_k(\cdot)$ can be avoided \cite{esgvi_paper},
\begin{subequations}
    \begin{align}
    \label{eq:gvi_qk_dx}
    &\expect{q_k}{\pd{}{\mbf{x}_k^{\trans}}\phi_k (\mbf{x}_k)} = \mbs{\Sigma}\inv_{kk} \expect{q_k}{(\mbf{x}_k - \mbs{\mu}_k)\phi_k(\mbf{x}_k)}, \\ 
    \label{eq:gvi_qk_ddx}
    &\expect{q_k}{\pd{^2}{\mbf{x}_k^{\trans} \p\mbf{x}_k}\phi_k (\mbf{x}_k)} = - \mbs{\Sigma}\inv_{kk}\expect{q_k}{\phi_k(\mbf{x}_k)} \\
    &\qquad + \mbs{\Sigma}\inv_{kk} \expect{q_k}{(\mbf{x}_k - \mbs{\mu}_k)(\mbf{x}_k - \mbs{\mu}_k)^{\trans}\phi_k(\mbf{x}_k)}\mbs{\Sigma}\inv_{kk} \nonumber \, .
\end{align}
\end{subequations}
Lastly, these marginal expectations can be  approximated using a cubature method \cite[\S6.3]{Sarkka}, that is
\begin{align}
    \label{eq:expectation_eval}
    \expect{q_k}{\phi_k(\mbf{x}_k)} \approx \sum_{l=1}^{L} w_{k,l}\phi_k(\mbf{x}_{k,l}),
\end{align}
where $w_{k,l}$ are the weights, and $\mbf{x}_{k,l}$ are the sigma points. The remaining expectations in~\eqref{eq:gvi_qk_dx} and~\eqref{eq:gvi_qk_ddx} are approximated in the same manner.

\subsection{ESGVI with Matrix Lie Group States}

Up to this point, the discussion has been restricted to vector-valued states $\mbf{x}_k$, for simplicity. However, as introduced previously, some robotics states are more naturally expressed as MLG states $\mbf{X}_k$. As such the variational algorithm~\eqref{eq:gvi_vector_update} must be reformulated to work with MLG states. To do so, uncertainty must be defined in a way that is compatible with the underlying group structure~\cite{Barfoot_Furgale_2014}. A Gaussian distribution associated with a MLG state is defined by perturbing the mean element $\mbfbar{X}$ with a zero-mean Gaussian random variable $\delta \mbs{\xi}$. Formally,
\begin{align}
    \label{eq:gaussian:mlg}
    \mbf{X} = \mbfbar{X} \oplus \delta \mbs{\xi}, \quad \delta \mbs{\xi} \sim \mc{N}(\mbf{0}, \mbs{\Sigma}),
\end{align}
where the $\oplus$ notation is borrowed from \cite{Solà_Deray_Atchuthan_2021}. A concise overview of the required MLG operations, including the definitions of $\oplus$ and its inverse $\ominus$, can be found in Appendix~\ref{sec:appendix:lie}. Having defined distributions associated with MLGs, the computational structure of the ESGVI algorithm can be reformulated for MLG-valued states. Although Stein's lemma was originally used in the Euclidean setting to rewrite the gradients of~\eqref{eq:gvi_qk_dx}, and~\eqref{eq:gvi_qk_ddx} in terms of expectations, its application remains valid when considering a perturbation $\delta\mbs{\xi}$ relative to $\mbfbar{X}$, as in~\eqref{eq:gaussian:mlg}. Specifically, computations are performed relative to the mean state~$\mbfbar{X}_k$ using~$\delta \mbs{\xi}_k$, which is Euclidean and therefore admits the same calculus rules as the standard vector space. In this way, the information update in~\eqref{eq:gvi_update_vector_1} and the gradient computation in~\eqref{eq:gvi_update_vector_2} proceed as before. The marginal MLG normal distribution parametrizes uncertainty through $\delta\mbs{\xi}_k$, meaning $q_k(\delta\mbs{\xi}_k)$, preserving the expectation formulation. Additionally, the Euclidean difference $(\mbf{x}_k - \mbs{\mu}_k)$, is replaced with $(\mbf{X}_k \ominus \mbfbar{X}_k)$. The expectations in~\eqref{eq:gvi_qk_dx}, and~\eqref{eq:gvi_qk_ddx} are then
\begin{subequations}
    \label{eq:gvi_grad_group}
    \begin{align}
        \label{eq:gvi_qk_dx_group}
        &\expect{q_k}{\pd{\phi_k (\mbf{X}_k)}{\delta \mbs{\xi}_k^{\trans}}} = \mbs{\Sigma}\inv_{kk} \expect{q_k}{(\mbf{X}_k \ominus \mbfbar{X}_k)\phi_k(\mbf{X}_k)}, \\ 
        \label{eq:gvi_qk_ddx_group}
        &\expect{q_k}{\pd{^2 \phi_k (\mbf{X}_k)}{\delta \mbs{\xi}_k^{\trans}\p \delta \mbs{\xi}_k}} = - \mbs{\Sigma}\inv_{kk}\expect{q_k}{\phi_k(\mbf{X}_k)} \\
        &\qquad + \mbs{\Sigma}\inv_{kk} \expect{q_k}{(\mbf{X}_k \ominus \mbfbar{X}_k)(\mbf{X}_k \ominus \mbfbar{X}_k)^{\trans}\phi_k(\mbf{X}_k)}\mbs{\Sigma}\inv_{kk} \nonumber \, .
    \end{align}
\end{subequations}
Although the expressions appear structurally similar to their Euclidean counterparts, their interpretation on MLGs requires further clarification, provided in Appendix~\ref{sec:appendix:stein_mlg}. Numerically, the expectations of~\eqref{eq:gvi_grad_group} are evaluated using a cubature method~\cite[\S6.3]{Sarkka} such that
\begin{subequations} \label{eq:mlg_cubatures}
    \begin{align}
        \expect{q_k}{\phi_k (\mbf{X}_k)} \!
            &\approx \! \sum_{\ell=1}^{L} w_k^{\ell} \, \phi_k(\mbc{X}_k^{\ell}), \\
        \expect{q_k}{(\mbf{X}_k \! \ominus \! \mbfbar{X}_k) \, \phi_k (\mbf{X}_k)} \!
            &\approx \! \sum_{\ell=1}^{L} w_k^{\ell} \, (\mbc{X}_k^{\ell} \ominus \mbfbar{X}_k) \, \phi_k(\mbc{X}_k^{\ell}),
    \end{align}
\end{subequations}
where the matrix-valued expectation associated with~\eqref{eq:gvi_qk_ddx_group} is omitted for brevity. The group sigma points $\mbc{X}_k^{\ell}$ are computed via~\eqref{eq:gaussian:mlg} as
\begin{align}
     \mbc{X}_k^{\ell} = \mbfbar{X}_{k} \oplus \delta \mbs{\xi}_{k}^{\ell},
\end{align}
where the perturbations are constructed as $\delta \mbs{\xi}_k^{\ell} = \sqrt{\mbs{\Sigma}_{kk}} \, \mbs{\alpha}^{\ell}$. Here, the unit sigma points $\mbs{\alpha}^{\ell}$, and their corresponding weights $w^{\ell}$, are generated by the same cubature method used in the Euclidean case of~\eqref{eq:expectation_eval}. Now that~\eqref{eq:gvi_qk_dx_group}, and~\eqref{eq:gvi_qk_ddx_group} have been generalized for the MLG state, the reformulation of~\eqref{eq:gvi_update_vector_3} follows naturally from~\eqref{eq:gaussian:mlg}. Instead of updating the Euclidean mean with a vector addition, the MLG mean $\mbfbar{X}^{(i)}$, is updated by mapping $\delta \mbs{\mu}$ back to the group using the exponential map, 
    \begin{equation}
    \mbfbar{X}^{(i+1)} = \mbfbar{X}^{(i)} \oplus \delta\mbs{\mu}.
\end{equation}
Here, the $\oplus$ notation is slightly abused, as each component $\mbfbar{X}_k^{(i)}$ is updated by the projected increment $\mbf{P}_k \delta \mbs{\mu}$, rather than the batch vector. This update ensures the new mean, $\mbfbar{X}_k^{(i+1)}$ remains in the group, while incorporating the computed variational variables, $\mbs{\Sigma}_{kk}\inv$ and $\delta\mbs{\mu}_k$. 
Recall from the earlier discussion in Section~\ref{sec:state_estimation:model} that the assumed form of $\phi_{k,\textrm{meas}}(\cdot)$ was based on the Skew-Laplace density function, 
\begin{align}
    \phi_{k, \textrm{meas}}(\mbf{X}_k) &= -\ln p(y_k | \mbf{X}_k)\nonumber, \\ &= - \left(\lambda \frac{y_k - g(\mbf{X}_k)}{\sigma^2} - \alpha \left|\frac{y_k - g(\mbf{X}_k)}{\sigma}\right|\right),
\end{align}
where normalization constants are safely dropped due to the optimization requiring only derivatives of $\phi_k(\mbf{X}_k)$. Similarly, the Gaussian process model is represented as, 
\begin{align}
    &\phi_{k, \textrm{proc}}(\mbf{X}_k) = -\ln p(\mbf{X}_k \mid \mbf{X}_{k-1}, \mbf{u}_{k-1})\nonumber, \\ 
    & \, = \onehalf (\mbf{X}_{k} \ominus \mbf{f}(\mbf{X}_{k-1}, \mbf{u}_{k-1}))^\trans \mbf{Q}_{k-1}^{-1} (\mbf{X}_{k} \ominus \mbf{f}(\mbf{X}_{k-1}, \mbf{u}_{k-1})).
\end{align}
The flexibility of $\phi_k(\mbf{X}_k)$ to represent any likelihood model, including Huber or Cauchy, was noted in~\cite{esgvi_paper}, and such factors are incorporated into ESGVI using the same procedure described above. Bringing these components together, the generalized form of~\eqref{eq:gvi_vector_update} for MLGs is reformulated as, 
\begin{subequations}
    \label{eq:gvi_group_update}
    \begin{align}
        \label{eq:gvi_update_group_1}
        \left(\mbs{\Sigma}\inv\right)^{(i+1)} &= \expect{q^{(i)}}{\pd{^2 \phi (\mbf{X})}{\delta \mbs{\xi}^{\trans}\p \delta \mbs{\xi}}}, \\
        \label{eq:gvi_update_group_2}
        \left(\mbs{\Sigma}^{-1}\right)^{(i+1)}\delta \mbs{\mu} &= - \expect{q^{(i)}}{\pd{\phi (\mbf{X})}{\delta \mbs{\xi}^{\trans}}}, \\
        \label{eq:gvi_update_group_3}
        \mbfbar{X}^{(i+1)} &= \mbfbar{X}^{(i)} \oplus \delta \mbs{\mu},
    \end{align}
\end{subequations}
where the gradients correspond to the expectations defined in~\eqref{eq:gvi_grad_group}. This completes the extension of the ESGVI update rules to MLG states. 

\section{Simulation and Experimental Results}
\label{sec:results}
The proposed method is validated in simulation and in experiments, and compared against two robust on-manifold MAP baselines, one employing the asymmetric Cauchy loss (MAP-C) shown in~\eqref{eq:asym_cauchy_pdf} and the other a GMM (MAP-GMM) to model skewed measurements. All estimators are implemented in \textsc{Python} within a common factor-graph-based framework. No comparison is made against a vector-valued ESGVI formulation, as the advantages of MLG-based state representations, such as improved consistency and avoidance of singularities, are well established in the literature~\cite{long2013banana, Barfoot_Furgale_2014}. The following sections present results first on simulated data, which allows controlled evaluation of performance, and then on real-world experiments to demonstrate practical applicability.

\subsection{Simulation Results}
\label{sec:sim_results}
The considered simulation comprises a planar robot equipped with wheel odometry, a gyroscope, and multiple range sensors capable of ranging to fixed landmarks of known position. The robot’s motion in discrete-time is modeled as 
%
\begin{align}
    \label{eq:proc_model}
    \mbf{X}_k \! = \! \mbf{X}_{k\-1} \oplus_r \! \left( \Delta t (\mbf{u}_{k\-1} \! + \! \mbf{w}_{k\-1}) \right), \quad
    \mbf{w}_{k\-1} \! \sim \! \mc{N}(\mbf{0}, \mbf{Q}_{k\-1}),
\end{align}
where $\mbf{X}_k \in \SE{2}$ denotes the robot pose at time step $k$. 
The input $\mbf{u}_{k-1}$ is a stacked vector of angular and linear velocity measurements obtained from the gyroscope and wheel odometry. Since these measurements are expressed in the robot’s body frame, the operator $\oplus_r$, is used to update the pose accordingly. Process noise $\mbf{w}_{k-1}$, is modeled as zero-mean Gaussian with covariance $\mbf{Q}_{k-1}$, and accounts for uncertainty in the motion estimate. To better reflect unmodeled dynamics such as wheel slip, the lateral velocity noise component is deliberately inflated. The robot's environment is cluttered, creating an alternating LOS and NLOS scenario. These skewed range measurements are generated as
\begin{align}
    \label{eq:meas_model}
    y_{k} &= \norm{\mbf{r}^{\ell_i w}_a - \mbf{C}_{ab}\mbf{r}^{p_j z}_b - \mbf{r}^{zw}_a}_2 + \nu_k, \quad
    \nu_k \sim \mc{D},
\end{align}
where $\mbf{r}^{\ell_i w}_a$ denotes the $i$th landmark position relative to point $w$ resolved in the absolute frame $\rframe{a}$, $\mbf{C}_{ab} \in \SO{2}$ is the rotation matrix mapping the body frame $\mc{F}_b$ to the absolute frame $\rframe{a}$, $\mbf{r}^{p_j z}_b$ is the known position of the $j$th range sensor relative to the gyroscope at $z$ resolved in the robot's body $\rframe{b}$, and similarly $\mbf{r}^{zw}_a$ is the position of the robot's gyroscope relative to the origin of the absolute frame, resolved in $\rframe{a}$. The skewed distribution $\mc{D}$ is generated by sampling from a zero-mean Gaussian, from which 25\% of measurements are additively perturbed by values drawn uniformly between $1\sigma$ and $6\sigma$, where the resulting distribution is illustrated in Figure~\ref{fig:uwb_fit_sim}. The objective is to estimate the full trajectory $\mbf{X} = \{ \mbf{X}_0, \ldots, \mbf{X}_K\}$. The proposed ESGVI algorithm is compared against two robust MAP-based approaches, one using an asymmetric Cauchy loss (MAP-C) and the other a GMM (MAP-GMM) of range measurement noise. Both are formulated to be compatible with MLG states for a fair comparison. Performance is evaluated using root-mean-squared error~(RMSE),
\begin{align}
    \label{eq:rmse}
    \textrm{RMSE} = \sqrt{\frac{1}{NK}\sum_{i=1}^N \sum_{k=1}^K\norm{\mbfhat{X}_{i,k} \ominus \mbf{X}_{i,k}}_2^2},
\end{align}
where $\mbfhat{X}_{i,k}$, and $\mbf{X}_{i,k}$ are the estimated and ground-truth states, for trajectory $i$, at time $t_k$, with RMSE reported separately for rotation~($\si{rad}$) and translation~($\si{m}$). Consistency, meaning the ability of the estimator to correctly quantify its own uncertainty, is measured using the average normalized estimation error squared~(aNEES)~\cite[\S5.4]{bar2001estimation},
\begin{align}
    \label{eq:anees}
    \textrm{aNEES} = \frac{1}{N Kn_x} \sum_{i=1}^{N}\sum_{k=1}^{K}(\mbfhat{X}_{i,k} \ominus \mbf{X}_{i,k})^{\trans} \mbs{\Sigma}_{i, kk}\inv (\mbfhat{X}_{i,k} \ominus \mbf{X}_{i,k}),
\end{align}
where a perfectly consistent estimator has an aNEES equal to one, due to the normalization in~\eqref{eq:anees} by the dimension of the state $n_x$. Performance comparisons are based on $50$~Monte Carlo trials over a trajectory of $400$~poses, with measurements and initial conditions regenerated randomly in each trial, with odometry being produced at $100$~($\si{Hz}$), and range measurements occurring at $10$~($\si{Hz}$). In all trials, the same GMM and Skew-Laplace parameters are used, fit using $5{,}000$~samples as shown in Figure~\ref{fig:uwb_fit_sim}. All algorithms are initialized with dead-reckoned gyroscope and wheel odometry measurements, though ESGVI can optionally be ``warm-started'' using a MAP-based solution to accelerate convergence. Finally, throughout this section, ESGVI uses a third-order Gauss-Hermite cubature method~\cite[\S6.3]{Sarkka} to evaluate the expectations in~\eqref{eq:mlg_cubatures}. As seen in Table~\ref{tab:sim_results_table}, all three estimators have similar performance. The GMM model is the most consistent, while the RMSE is comparable between the ESGVI and the MAP estimators. ESGVI is slightly overconfident compared to the MAP estimators, which follows the results of~\cite{Bishop_2006} as minimizing the KL-divergence, $\textrm{KL}(q||p)$ can result in a Gaussian that is too confident, underestimating the variance.
\begin{table}
\centering
\caption{Estimation performance over 50 Monte Carlo trials.}
\label{tab:sim_results_table}
\resizebox{\linewidth}{!}{
\begin{tabular}{lccc}
    \toprule
    \textbf{Method} & \textbf{RMSE ($\si{rad}$)} & \textbf{RMSE ($\si{m}$)} & \textbf{aNEES} \\
    \midrule
    MAP-C & 0.026 & 0.050 & 0.916 \\
    MAP-GMM   & 0.026 & \textbf{0.046} & \textbf{0.990} \\
    ESGVI        & \textbf{0.025} & \textbf{0.046} & 1.019 \\
    \bottomrule
\end{tabular}
}
\vspace{-1.5em}
\end{table}
\subsection{Experimental Setup}
\begin{figure}[htbp!]
    \centering
    \includegraphics[width=\columnwidth]{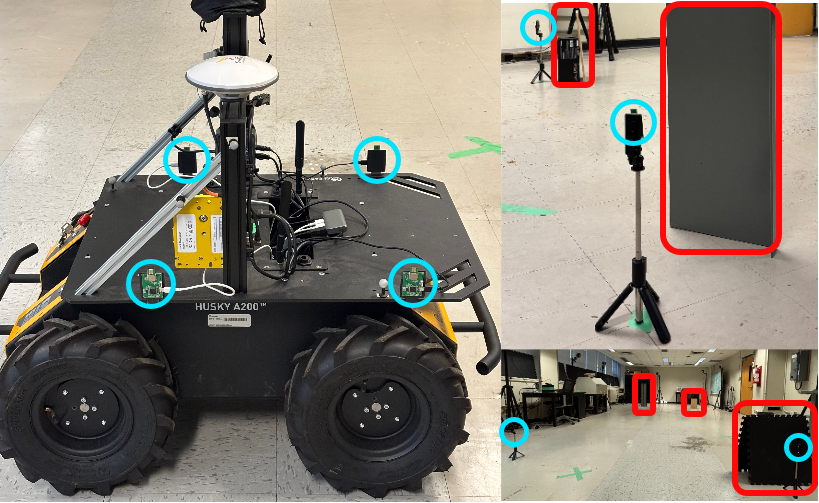}
    \caption{Experimental setup showing the Clearpath Husky platform equipped with four UWB tags operating in a cluttered environment with four fixed UWB anchors. UWB modules are circled in blue, and obstacles are outlined in red.}
    \label{fig:experimental_setup}
    \vspace{-0.5em}
\end{figure}
\begin{figure}[htbp!]
    \centering
    \includegraphics[width=\columnwidth]{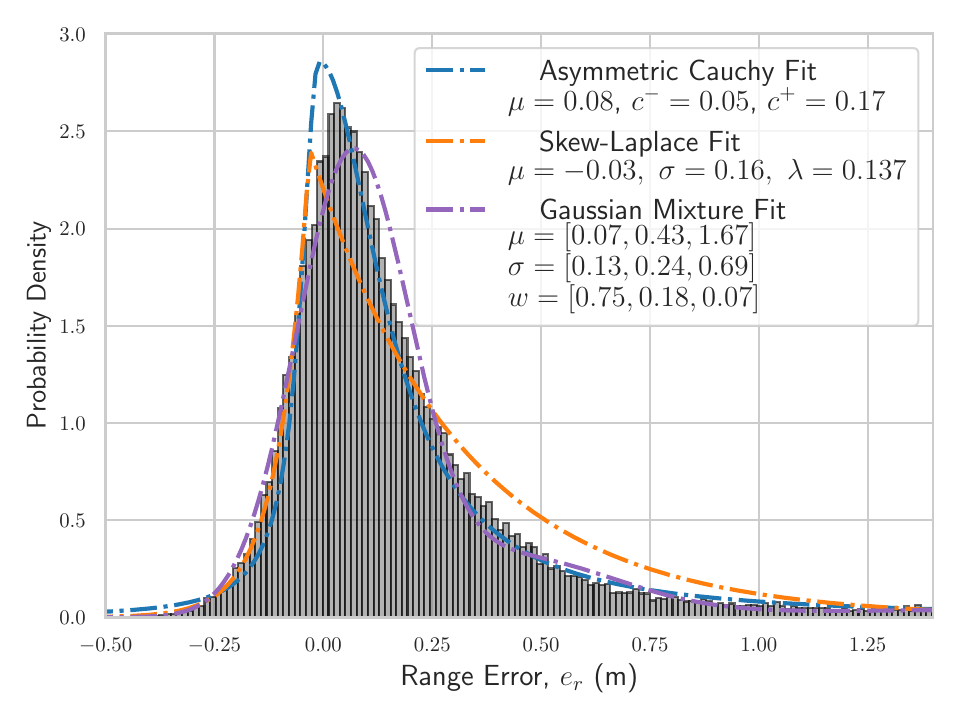} 
    \caption{Calibrated UWB range error histogram across all trajectories with fitted noise models.}
    \vspace{-1.5em}
    \label{fig:uwb_histogram}
\end{figure}

Experiments were conducted using the Clearpath Husky vehicle, equipped with four DWM1000 Decawave UWB tags, as shown in Figure~\ref{fig:experimental_setup}. These tags measure ranges to a set of fixed UWB anchors positioned throughout the environment, which are then used for localization. UWB ranging relies on measuring the time-of-flight of radio signals \cite{Shalaby_Cossette_Forbes_Le_Ny_2023} and is therefore sensitive to non-line-of-sight (NLOS) propagation and multipath effects, which can introduce bias in the measured ranges, as illustrated in Figure~\ref{fig:problem}. Such conditions are common in indoor environments, where obstacles, clutter, and frequent foot traffic create dynamic obstructions that lead to skewed measurements. To reproduce these effects in a controlled manner, obstacles made of metal, wood, and foam were placed near selected UWB anchors, as shown in Figure~\ref{fig:experimental_setup}.
\begin{figure}[htbp!]
    \centering
    \vspace{-1.4em}
    \includegraphics[width=\columnwidth]{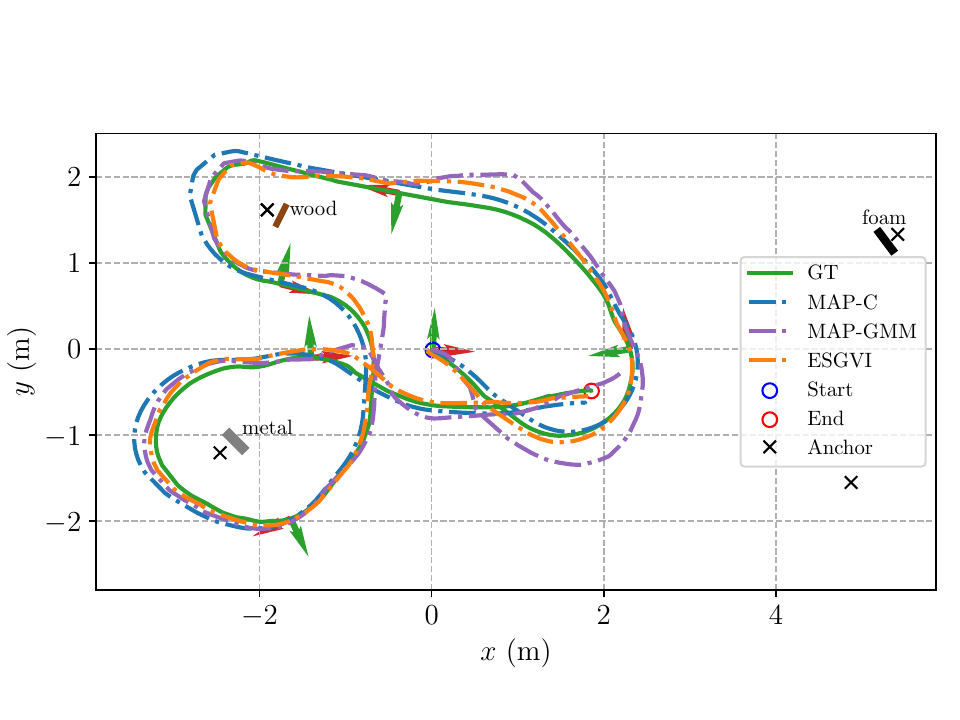}
    \vspace{-2em}
    \caption{Example trajectory (Trajectory 6) driven by the Husky through a cluttered environment, comparing estimator outputs against ground-truth data recorded by a motion capture system.}
    \vspace{-1.5em}
    \label{fig:mocap_traj_cluttered}
\end{figure}
The Clearpath Husky was driven along ten independent trajectories designed to randomly expose the UWB range measurements to alternating LOS and NLOS conditions. An example of the sixth trajectory is shown in Figure~\ref{fig:mocap_traj_cluttered}. Ground-truth poses were recorded at $120$~(\si{Hz}) using a Vicon motion capture system and used for estimator performance evaluation. The robot was operated at a nominal forward velocity of $0.4$~(\si{m/s}), resulting in trajectories of approximately $60$~seconds each in duration within a $40$~(\si{m^2}) indoor test area. UWB range measurements were collected at $30$~(\si{Hz}) per tag, yielding over $75{,}000$ range observations across all ten trajectories. 
In addition to NLOS effects, UWB measurements are affected by other error sources, including antenna delays, irregular radiation patterns, and clock skews between UWB tags. To account for these factors, the calibration procedure described in \cite{Shalaby_Cossette_Forbes_Le_Ny_2023} was applied. In particular, the antenna-delay calibration was performed, and the data-driven model relating received signal power to attitude-dependent range bias was reused to isolate for NLOS-induced bias effects. Range measurement errors were then computed with respect to the ground-truth poses, across all trajectories. The resulting error distribution, shown in Figure~\ref{fig:uwb_histogram}, clearly demonstrates the influence of obstacles through its pronounced skewed tails. Based on this distribution, the asymmetric Cauchy, GMM, and Skew-Laplace models were fitted. The GMM utilized three mixture components, consistent with the methodology of \cite{Zhao_Goudar_Tang_Qiao_Schoellig_2023}, where UWB errors were modeled in a similar fashion. All three models effectively capture the skewness arising from the prevalence of NLOS measurements, with the Skew-Laplace model exhibiting the most conservative behavior for highly biased ranges, while the asymmetric Cauchy and GMM models more closely align with the empirical histogram. Notably, the experimental error distribution closely mirrors the simulated case shown in Figure~\ref{fig:uwb_fit_sim}, reinforcing the validity of the simulation assumptions.

\subsection{Experimental Results}
\begin{figure*}
    \vspace{-0.5em}
    \centering
    \includegraphics[width=1\linewidth]{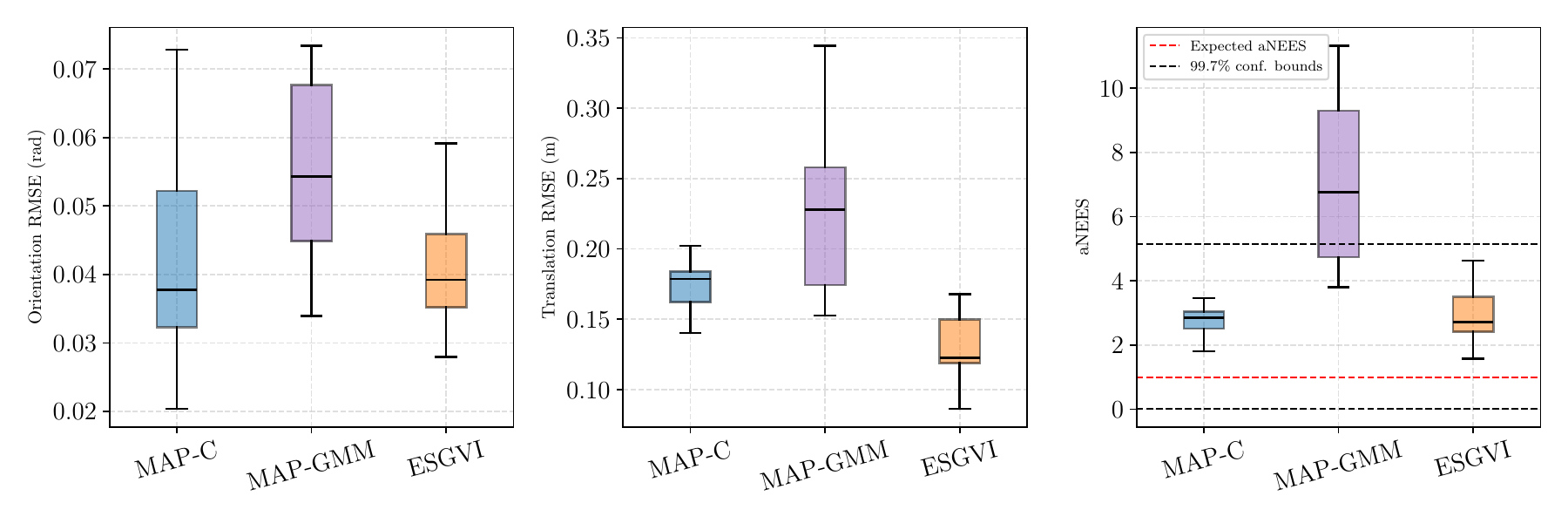}
    \caption{Distribution of orientation RMSE, translation RMSE, and aNEES across 10 experimental trajectories for each estimator. The consistency bounds for the aNEES are computed using the median trajectory length.}
    \label{fig:boxplot_exp}
    \vspace{-1.5em}
\end{figure*}

The experimental evaluation follows the same process and measurement models used in the simulation study presented in Section~\ref{sec:sim_results}. Specifically, the process dynamics remain as defined in~\eqref{eq:proc_model}, while the UWB measurement model follows~\eqref{eq:meas_model}, with parameters identified from the empirical error distributions aggregated across all ten trajectories shown in Figure~\ref{fig:uwb_histogram}. Each estimator was initialized using dead-reckoned gyroscope and wheel odometry measurements prior to evaluation across all cluttered trajectories. The results, summarized in Figure~\ref{fig:boxplot_exp}, demonstrate that ESGVI provides improved positional accuracy and comparable consistency and orientation accuracy relative to the asymmetric Cauchy MAP estimator. Notably, unlike in the simulation study, the GMM-based estimator performs less effectively in the experimental setting. This degradation is likely due to varying proportions of NLOS measurements across trajectories, which hinders the GMM’s ability to generalize, in contrast to the Monte Carlo simulations where NLOS measurements occurred at a fixed rate of 25\%. From Figure~\ref{fig:mocap_traj_cluttered}, it is evident that estimator performance deteriorates when the vehicle traverses NLOS regions caused by the metal, while other obstacles appear to have a less pronounced effect on the trajectory. On the other hand, both the asymmetric Cauchy MAP and ESGVI estimators exhibit comparable orientation RMSE, consistent with the findings in~\cite{esgvi_paper}, which demonstrated that the performance advantages of the vector-valued ESGVI algorithm over MAP methods are primarily in translational accuracy rather than orientation estimation. This trend appears to extend to the MLG case as well. Regarding consistency, all estimators tend to be slightly overconfident, with the asymmetric Cauchy MAP and ESGVI exhibiting the most similar behavior. This overconfidence is expected because the metric is computed separately for each independent trajectory, and the proportion of NLOS measurements varies from one trajectory to another, making it a challenging dataset.

\subsection{Limitations}

Although ESGVI demonstrates improved accuracy, it is computationally more expensive than MAP-based methods, often an order of magnitude slower in practice. As noted in~\cite{esgvi_paper}, ESGVI has the same theoretical asymptotic complexity as MAP, since both require a lower–diagonal–upper decomposition of the information matrix. However, the cubature-based computation of marginal expectations increases the constant factor, resulting in higher runtimes. In practice, this is further exacerbated by the lack of speed-oriented optimizations and the use of a simple backtracking line search.

While warm-starting ESGVI from a MAP solution improves convergence, the current implementation remains restricted to offline use and is therefore unsuitable for real-time applications. For a quantitative runtime evaluation, please refer to the relevant section of~\cite{esgvi_paper}. Nevertheless, ESGVI remains valuable in scenarios where robustness and improved estimation performance outweigh computational cost, and accelerating ESGVI remains an open direction for future work.

    
    
    

\section{Conclusion}
\label{sec:conclusion}
In this letter, a more flexible and MLG compatible ESGVI method was presented. Building on the work of~\cite{esgvi_paper}, it was shown how ESGVI can naturally incorporate skewed measurement noise models while supporting states defined on MLGs. The method was evaluated in an outlier-rich UWB localization experiment, demonstrating greater accuracy and comparable consistency relative to on-manifold robust MAP solutions using GMMs or asymmetric Cauchy losses to handle outliers. Although experimentally ESGVI was more computationally intensive than MAP, these results highlight its potential as a robust and flexible framework for state estimation in challenging environments.

\section*{Acknowledgments}
The authors would like to thank Kyle Biron-Gricken and Nicholas Dahdah for their assistance with the experimental setup, and Vassili Korotkine for helpful discussions on GMMs.

\printbibliography

\appendix
\raggedbottom
\section{Appendix} \label{sec:appendix}

\subsection{Matrix Lie Groups} \label{sec:appendix:lie}

Matrix Lie groups (MLGs) consist of elements that are matrices, typically denoted as $\mbf{X} \in G$. To relate the Lie algebra, $\mathfrak{g}$, with its associated vector space, $\rnums^g$, two key mappings are defined: the \emph{Hat} operator, which maps a vector $\mbs{\tau} \in \rnums^g$ to its Lie algebra element $\mbs{\tau}^\wedge \in \mathfrak{g}$, and the \emph{Vee} operator, which performs the inverse mapping from $\mbs{\tau}^\wedge$ back to $\mbs{\tau}$, given by
\begin{align}
    \textrm{Hat}:& \quad \rnums^g \rightarrow \mathfrak{g}; \quad \mbs{\tau} \mapsto \mbs{\tau}^\wedge, \\
    \textrm{Vee}:& \quad \mathfrak{g} \rightarrow \rnums^g; \quad \mbs{\tau}^\wedge \mapsto (\mbs{\tau}^\wedge)^\vee = \mbs{\tau}.
\end{align}
The exponential and logarithmic maps describe the relationship between the Lie algebra $\mathfrak{g}$, and the MLG $G$, and are defined as
\begin{align}
    \exp:& \quad \mathfrak{g} \rightarrow G; \quad \mbs{\tau}^\wedge \mapsto \mbf{X} = \exp(\mbs{\tau}^\wedge), \\
    \log:& \quad G \rightarrow \mathfrak{g}; \quad \mbf{X} \mapsto \mbs{\tau}^\wedge = \log(\mbf{X}).
\end{align}
The plus and minus operators are a concise means to describe operations involving elements of the MLG $G$ and the Lie algebra $\mathfrak{g}$. Given that the composition of two group elements is not commutative, there are right and left versions of the plus and minus operators. The right versions are
\begin{align}
    \label{eq:right_oplus}
    \oplus_r&: \, \mbf{Y} = \mbf{X} \oplus_r \delta \mbs{\xi} \triangleq \mbf{X}\exp(\delta \mbs{\xi}^{\wedge}), \\
    \label{eq:right_ominus}
    \ominus_{r}&:\, \mbs{\tau} = \mbf{Y} \ominus_r \mbf{X} \triangleq \log(\mbf{X}\inv \mbf{Y})^\vee,
\end{align}
and the left operations are
\begin{align}
    \label{eq:left_oplus}
    \oplus_\ell&: \mbf{Y} = \mbf{X} \oplus_\ell \delta \mbs{\xi} \triangleq \exp(\delta \mbs{\xi}^{\wedge})\mbf{X}, \\
    \label{eq:left_ominus}
    \ominus_{\ell}&: \mbs{\tau} = \mbf{Y} \ominus_\ell \mbf{X} \triangleq \log( \mbf{Y}\mbf{X}\inv)^\vee .
\end{align}
If no subscript is indicated when writing either $\oplus$ or $\ominus$, it is assumed that the expression holds for both directions, as long as the chosen direction remains consistent. For a more detailed review of matrix Lie groups, please refer to \cite{Solà_Deray_Atchuthan_2021}.

\subsection{Stein's Lemma for Matrix Lie Groups}
\label{sec:appendix:stein_mlg}
The expressions in~\eqref{eq:gvi_grad_group} are not new results but rather simply Stein’s identity written in Lie-group notation for the marginal MLG density. For the perturbation variable $\delta\mbs{\xi}_k \sim \mathcal{N}(\mbf{0},\mbs{\Sigma}_{kk})$ and any smooth test function 
$f(\delta\mbs{\xi}_k) = \phi_k(\bar{\mbf{X}}_k \oplus \delta\mbs{\xi}_k)$, Stein’s lemma actually takes the form
\begin{subequations}\label{eq:stein_true}
\begin{align}
    \expect{}{\tfrac{\partial f}{\partial \delta \mbs{\xi}_k^\trans}}
        &= \mbs{\Sigma}_{kk}^{-1}\,\expect{}{\delta \mbs{\xi}_k\, f(\delta \mbs{\xi}_k)}, \\
    \expect{}{\tfrac{\partial^2 f}{\partial \delta \mbs{\xi}_k\,\partial \delta \mbs{\xi}_k^\trans}}
        &= -\mbs{\Sigma}_{kk}^{-1}\expect{}{f(\delta \mbs{\xi}_k)}\nonumber \\
        & \qquad \quad
           + \mbs{\Sigma}_{kk}^{-1}\expect{}{\delta \mbs{\xi}_k \delta \mbs{\xi}_k^\trans f(\delta \mbs{\xi}_k)}\,\mbs{\Sigma}_{kk}^{-1}.
\end{align}
\end{subequations}
Using the retraction $\mbf{X}_k = \bar{\mbf{X}}_k \oplus \delta\mbs{\xi}_k$ and the corresponding difference $\mbf{X}_k \ominus \bar{\mbf{X}}_k$, these relations can be rewritten in the Lie-group form shown in~\eqref{eq:gvi_grad_group}. This rewriting is exact when $\log$ and $\exp$ acts as a local inverse, which holds within a neighborhood of the identity.



\end{document}